\documentclass{llncs}
\pdfoutput=1

\usepackage{graphicx}
\usepackage{amssymb}
\usepackage{booktabs}

\title{Difficulty Rating of Sudoku Puzzles: An~Overview and Evaluation}  
  
\author{Radek Pel\'{a}nek}

\institute{Faculty of Informatics, Masaryk University Brno}

\begin{document}

  \maketitle

  \begin{abstract}
    How can we predict the difficulty of a Sudoku puzzle? We give an overview
    of difficulty rating metrics and evaluate them on extensive dataset on
    human problem solving (more then 1700 Sudoku puzzles, hundreds of solvers).
    The best results are obtained using a computational model of human solving
    activity. Using the model we show that there are two sources of the problem
    difficulty: complexity of individual steps (logic operations) and structure
    of dependency among steps. We also describe metrics based on analysis of
    solutions under relaxed constraints -- a novel approach inspired by phase
    transition phenomenon in the graph coloring problem. In our discussion we
    focus not just on the performance of individual metrics on the Sudoku
    puzzle, but also on their generalizability and applicability to other
    problems.
  \end{abstract}


\section{Introduction}
\label{sec:introduction}

What determines difficulty of problems for humans? Is there any relation
between difficulty for humans and computational complexity for computers? Can
we predict difficulty of problems algorithmically? Answers to these questions
can give us an insight into human cognition and provide impulses for
development of artificial intelligence. Moreover, these questions have
important practical applications in human-computer collaboration and in
training of problem solving skills, e.g., for developing intelligent tutoring
systems~\cite{anderson1985intelligent,beck1997using,caine2007tutoring}.

Constraint satisfaction problems (CSP) are well suited for this kind of
research, because many popular puzzles have natural CSP formulation. This is
particularly the case of Sudoku puzzle, which is at the same time a ``textbook
example'' of CSP and very popular puzzle.

\subsection{Human Problem Solving and Problem Difficulty}

Although human problem solving has been studied for at least 40
years~\cite{simon1972hps}, the issue of problem difficulty has not been
properly explored. Results are available only for several specific puzzles
(mostly transport puzzles): Tower of Hanoi (and its
izomorphs)~\cite{kotovsky1985some}, Chinese rings~\cite{kotovsky1990msp},
15-puzzle~\cite{pizlo2005solving}, traveling salesman
problem~\cite{dry2006human}, Sokoban~\cite{sokoban-stairs}, and Rush hour
puzzle~\cite{transport-flairs}. The main reason for the scarceness of research
in this important area is probably the difficulty of obtaining sufficient data
for analysis of human problem solving. To estimate ``human difficulty'' of a
problem we need many people to solve it; to study techniques for predicting
problem difficulty we need data on many problems; thus we need data from
hundreds of hours of problem solving. It is practically impossible to obtain
data of such extend using the traditional laboratory approach to data
collection.

However, the advent of ubiquitous Internet allows a scalable alternative to the
traditional approach to data collection. In our other
work~\cite{sokoban-stairs,transport-flairs}, we build our own web portal and
use it to collect data on human problem solving. In this work, we exploit the
popularity of the Sudoku puzzle and use data from several web servers. In this
way we have been able to obtain (rather easily and quickly) data capturing
hundreds of thousands of hours of human problem solving activity (approximately
2000 puzzles, hundreds of human solvers for each puzzle). This means that our
data are several orders of magnitude more extensive than the usual data used in
study of human problem solving -- most previous research uses data based on
tens or hundreds of hours of human problem solving activity (usually about 20
people and 5 puzzles). Even though this way of data collection has its
disadvantages (e.g., lack of direct control over participants), we show that
thanks to the scale of the ``experiment'', the data are robust and applicable
for research purposes.

In this work we study the problem difficulty of one particular problem -- the
Sudoku puzzle. We describe difficulty rating metrics for the Sudoku puzzle, and
evaluate their correlation with human performance (measured by time). This goal
has direct applications -- such metrics are heavily used since Sudoku is
currently very popular and even commercially important~\cite{sudoku-milion},
and the difficulty rating of puzzles is one of the key things which influence
the user's experience of puzzle solving. Despite the straightforwardness of our
goal and its direct applicability, there is no easily applicable theory that
could be used to guide the development of difficulty rating metrics. Currently
used Sudoku metrics are usually built in an ad hoc manner, they are not
properly evaluated and their merits are not clear.

Although we only analyze data on a single puzzle, the aim of this work goes
beyond the specific study of Sudoku. We would like to raise interest in the
study of problem difficulty, for example by showing that extensive and robust
data for study are easily available on the Internet. Our main goal is not to
get ``the best Sudoku rating metric'', but rather to get an insight into
problem difficulty, which could be applied also to other problems. Therefore,
we do not focus just on the performance of individual metrics, but also on
their generalizability and potential applicability to other problems.

\subsection{The Sudoku Puzzle}

Sudoku is a well-known number placement puzzle: for a partially filled $9\times
9$ grid, the goal is to place numbers 1 to 9 to each cell in such a way that in
each row, column, and $3\times 3$ sub-grid, each number occurs exactly once. We
focus on the Sudoku puzzle for several reasons. The main reason has already
been stated above -- thanks to the current popularity of Sudoku, we can easily
obtain large scale data on human solving activity. Moreover, the Sudoku puzzle
has very simple rules, which makes it suitable for extensive analysis. Finally,
Sudoku is also a member of an important class of constraint satisfaction
problems. The class of constraint satisfaction problems contains many other
puzzles and also many real life problems (e.g., timetabling, scheduling).
Although we use data for the Sudoku puzzle, our goal is to make the analysis
and difficulty metrics as general as possible, so that the results are
potentially applicable to other constraint satisfaction problems.

Sudoku has been the subject of many research studies, particularly with respect
to its mathematical and algorithmic properties, e.g., enumerating possible
Sudoku grid~\cite{felgenhauer2005enumerating}, NP-completeness of generalized
version of Sudoku~\cite{yato2003complexity}, formulation as constraint
satisfaction problem~\cite{simonis2005sudoku,lynce2006sudoku}, and application
of a large number of artificial intelligence and constraint satisfaction
techniques for solving and generating Sudoku puzzles
~\cite{mantere2007solving,o2007generating,sudoku-fowler,yue2006sudoku,lewis2007metaheuristics,nicolau2006solving,geem2010harmony,moraglio2008geometric,jilg2009sudoku,goldbergersolving,moon2006multiple}.
Also some psychological aspects of the puzzle have been
studied~\cite{lee2008psychological}.

The difficulty rating of Sudoku puzzles is, of course, not a novel problem. The
issue of the Sudoku difficulty rating is widely discussed among Sudoku players
and developers, but it has not been a subject of a serious scientific
evaluation. Current rating algorithms are based mainly on personal experiences
and ad hoc tuning. There are several research papers which discuss methods for
difficulty rating~\cite{simonis2005sudoku,mantere2007solving,henz2009sudoku};
however, these works study the correlation of proposed metric with the
difficulty rating provided by the puzzle source (usually a newspaper), not with
the data on human performance. Such analysis is not very meaningful since the
rating provided in puzzle sources is just another rating provided by a computer
program (nearly all published puzzles are generated and rated by a computer).
The only work that we are aware of and that uses data on real human performance
is the brief report by Leone et al.~\cite{sudoku-bagging}. We have reported on
preliminary versions of this research in~\cite{sudoku-flairs,sudoku-tr}.

\subsection{Contributions}

We provide an analysis of the used datasets on Sudoku solving, and using these
datasets we evaluate different approaches to the difficulty rating of Sudoku
puzzles.

The first class of metrics is based on direct modeling of human problem solving
of Sudoku puzzles, i.e., the use of ``logic techniques'' to solve the puzzle,
which corresponds to the constraints propagation approach in CSP solving. Here
we determine two main aspects of problem difficulty. The first is the
complexity of individual steps (logic operations) involved in solving the
problem -- this is the usual approach used for rating Sudoku puzzles (see e.g.,
\cite{sudoku-explainer,sudoku-fowler}). We show that there is also a second
aspect that has not yet been utilized for difficulty rating -- the structure of
dependency among individual steps, i.e., whether steps are independent (can be
applied in parallel) or whether they are dependent (must be applied
sequentially). We provide a simple general model that captures both of these
aspects. We show that even with rather simple rating metric with few
Sudoku-specific details we can obtain hight correlation with human performance
-- the Pearson's correlation coefficient is 0.88. By combination with
previously proposed Sudoku-specific metrics we improve the correlation to 0.95.

Metrics based on modeling human behaviour using constraint propagation give
very good prediction, but they are not easily applicable to other problems.
Sudoku has very simple rules and thus it is quite easy to formalize human
``logic techniques''. Even for similar and still rather simple problems (like a
Nurikabe puzzle), it can be quite difficult to formulate human logic techniques
(choose suitable constraint propagation rules) and thus obtain a model of human
problem solving and a good rating metric.

Therefore, we also study other types of metrics. We found an interesting
approach to difficulty rating, which is based on analysis of solutions under
relaxed constraints. This novel technique is inspired by phase transition
phenomenon in random constraint satisfaction problems (particularly in the
graph coloring
problem)~\cite{cheeseman1991really,monasson1999determining,krzakala2008phase}.
Metrics inspired by notions like freezing~\cite{krzakala2008phase} and backbone
variable~\cite{monasson1999determining} are slightly worse than metrics based on
the constraint propagation approach, but the constraint relaxation metrics is
much easier to transfer to other problems than the model-based approach (as we
illustrate on the case of Nurikabe puzzle) and it gives an interesting insight
into human problem solving -- it does not mimic human thinking and yet it can
predict it.

We also describe experiments with metrics based on other algorithmic approaches
to solve Sudoku (e.g., backtracking search, simulated annealing). However,
these metrics lead to poor predictions, sometimes even worse than the baseline
metric ``number of givens''.





\section{Background}

In this section we provide background information about constraint satisfaction
problems, the Sudoku puzzle, and datasets used for evaluation.


\subsection{Constraint Satisfaction Problems and the Sudoku Puzzle}

The Sudoku puzzle can be formulated as a constraint satisfaction problems. A
constraint satisfaction problem is given by a set of variables $X = \{ x_1,
\ldots, x_n \}$, a set of domains of variables $\{ D_1, \ldots, D_n \}$, and a
set of constraints $\{C_1, \ldots, C_m \}$. Each constraint involves some
subset of variables and specifies allowed combinations of variable values
(usually given in a symbolic form, e.g., $x_1 \neq x_2$). A solution of a
constraint satisfaction problem is an assignment of values to all variables
such that all constraints are satisfied. The class of CSPs contains many
puzzles (e.g., the eight queen problem, the cryptarithmetic puzzle) as well as
many important practical problems (map coloring problems, timetabling problems,
transportation scheduling).

Sudoku puzzle is a grid of $9\times 9$ cells, which are divided into nine
$3\times 3$ sub-grids, partially filled with numbers 1 to 9. The solution of
the puzzle is a complete assignment of numbers 1 to 9 to all cells in the grid
such that each row, column and sub-grid contains exactly one occurrence of each
number from the set $\{1, \ldots, 9\}$. Sudoku puzzle is well-posed, if it
admits exactly one solution. We study only well-posed puzzles.

\begin{figure}[tbp]
  \centering
  \includegraphics[width=\linewidth]{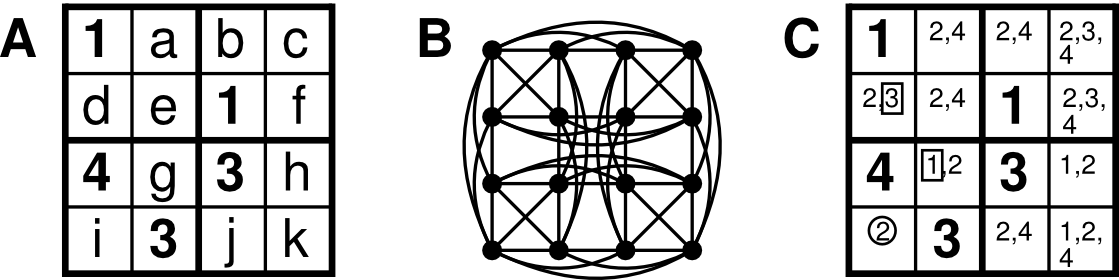}
  \caption{A) An example of a $4\times 4$ Sudoku puzzle, empty cells are
    denoted as variables. B) Sudoku graph for a $4\times 4$ puzzle. C) The
    puzzle with enumerated candidate sets. Circle marks naked single, rectangle
    marks hidden single.}
  \label{fig:sudoku4x4background}
\end{figure}

Sudoku puzzle can be easily generalized for any grid size of $n^2 \times n^2$
and values from 1 to $n^2$~\cite{simonis2005sudoku}. Moreover, there are many
variants of Sudoku which use non-regular sub-grids (e.g, pentomino), or
additional constraint (e.g., arithmetic comparison of values). In this work we
use $4 \times 4$ Sudoku puzzle (Fig.~\ref{fig:sudoku4x4background}) as a
small illustrating example, otherwise we consider solely the classical $9\times
9$ Sudoku puzzles.

Sudoku can be easily expressed as a constraint satisfaction
problem~\cite{simonis2005sudoku}. Consider an example of a $4\times 4$ Sudoku
given in Fig.~\ref{fig:sudoku4x4background}; each cell corresponds to one
variable, the domain of each variable is a set $\{1, 2, 3, 4\}$ and the
constraints express non-equality of variables and constants in the same row,
column or sub-grid, e.g., for variable $j$ we have the following constraints:
$j\neq b, j\neq h, j\neq i, j\neq k, j\neq 3,$ \mbox{$j\neq 1$}.

Sudoku can also be viewed as a special case of a graph coloring problem, which
is a particularly well-studied CSP. Given a graph $G$, the problem is to decide
whether its vertices can be colored by $k$ colors in such a way that no two
vertices connected by an edge share a color. For the Sudoku puzzle, we can
construct a ``Sudoku graph'' $G_S$ (see Fig.~\ref{fig:sudoku4x4background}).
The Sudoku puzzle thus becomes 9-colorability problem for the graph $G_S$ with
an initial color assignment\footnote{Note that it is also possible to reduce
  Sudoku to pure graph coloring problem by adding vertices corresponding to
  numbers 1 to 9.}. The connection between Sudoku and graph coloring is further
elaborated by~Herzberg and Murty~\cite{herzberg2007sudoku}.

\subsection{Solving Sudoku}
\label{sec:solving-sudoku}

Now we discuss algorithmic approaches to Sudoku solving and comment on their
relation to human solving.

\subsubsection{Backtracking}

\begin{figure}[tbp]
   \centering
   \includegraphics[width=\linewidth]{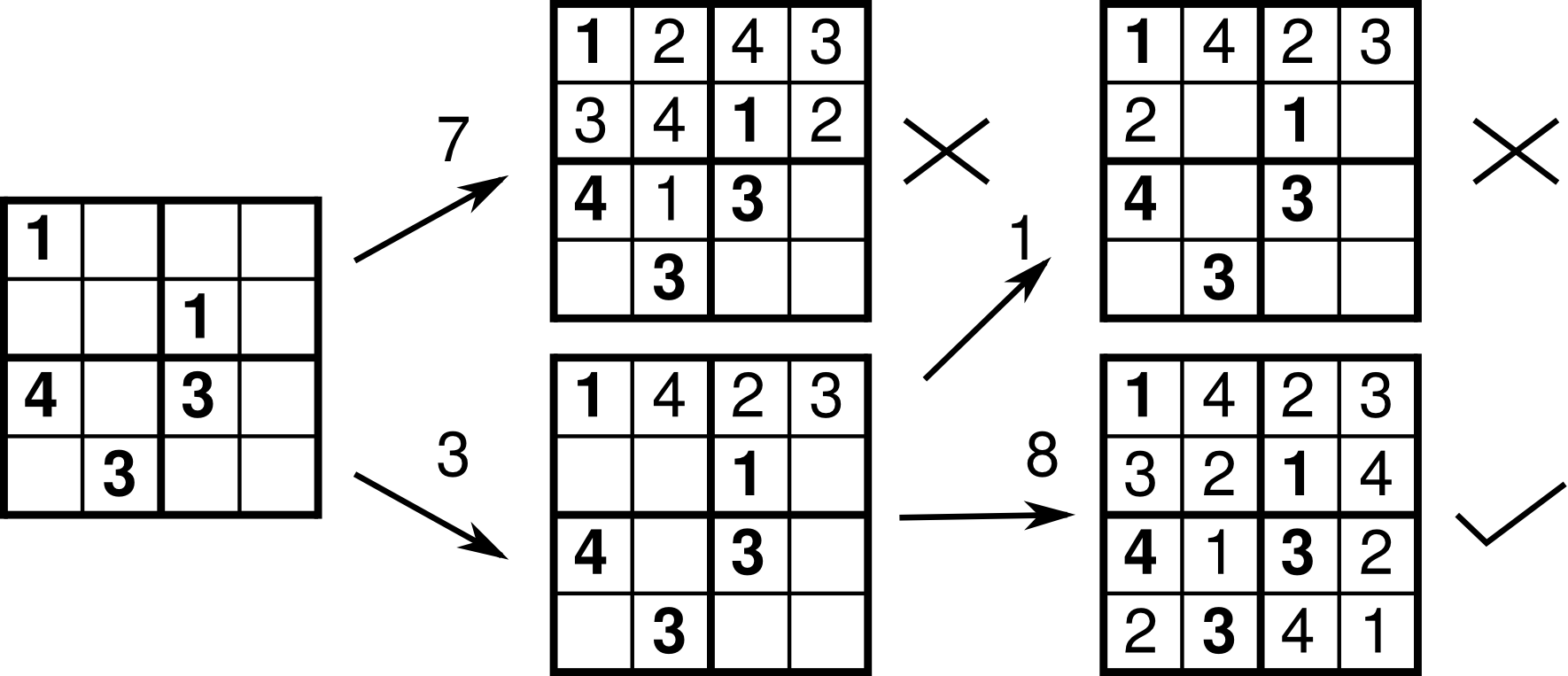}
   \caption{Comprised version of a search tree of a backtracking algorithm on a
     $4\times 4$ Sudoku puzzle. Numbers over arrows indicate number of steps
     (filled numbers) without branching.}
   \label{fig:sudoku4x4backtracking}
 \end{figure}

The basic approach to solving CSPs is the backtracking search. It starts with
an empty variable assignment and tries to find a solution by assigning values
to variables one by one. Whenever it finds a violation of a constraint, it
backtracks. In this way the search explores the tree of feasible partial
assignments. Fig.~\ref{fig:sudoku4x4backtracking} shows the search tree for a
sample $4\times 4$ Sudoku (for readability of the figure we select variables
``by rows'', a more efficient approach is to select the most constrained
variable in each step).

The run time of a backtracking algorithm grows exponentially with the number of
variables. Nevertheless, the $9\times 9$ Sudoku puzzle can be easily solved by
computer using the backtracking search. For humans, however, this is not a
favoured approach. As can be seen in Fig.~\ref{fig:sudoku4x4backtracking},
even for a small $4\times 4$ Sudoku, some branches in the search can be quite
long. For humans, systematic search is laborious, error-prone, and definitely
not entertaining.

\subsubsection{Constraint Propagation}
\label{sec:constr-prop}

Another approach to solving CSP is to find values of (some) variables by
reasoning about constraints. For each variable $x_i$ we define a current
candidate set -- a set of such values that do not lead to direct violation of
any constraint (see Fig.~\ref{fig:sudoku4x4background}). By reasoning about
candidate sets and constraints we can often derive solution without any search.

For our example from Fig.~\ref{fig:sudoku4x4background}, it is easy to see
that the value of variable $i$ must be $2$, because constraints restrict the
domain of the variable to a single value. Thus we can assign the value to the
variable without any search. Using this partial assignment we can straighten
constraints on other variables (and show, for example, that also $h$ has to be
$2$).

Constraint propagation is not guaranteed to find a solution, but it may be more
efficient than backtracking search and can also be combined with backtracking
search to produce superior results. We are interested in constraint propagation
particularly because this is the natural way how humans try to solve CSPs.

Let us consider specifically the case of the Sudoku puzzle. Human solving of
Sudoku proceeds by sequence of steps in which (correct) values are filled into
cells. Two basic techniques directly correspond to the rules of the puzzle (see
also Fig.~\ref{fig:sudoku4x4background}):
\begin{description}
\item[Naked single technique] (also called singleton, single value, forced
  value, exclusion principle): For a given cell there is only one value that
  can go into the cell, because all other values occur in row, column or
  sub-grid of the cell (any other number would lead to a direct violation of
  rules).
\item[Hidden single technique] (also called naked value, inclusion principle):
  For a given unit (row, column or sub-grid) there exists only one cell which
  can contain a given value (all other placements would lead to a direct
  violation of rules).
\end{description}

Sudoku problems solvable by iteration of these two techniques are further
denoted as ``simple Sudoku''. Most of the publicly used puzzles which are
ranked as easy or mild are simple Sudokus. There exists many advanced
techniques, such as pointing, chaining, naked and hidden pairs (see, e.g.,
Sudoku Explainer~\cite{sudoku-explainer}), but we do not elaborate on these
techniques in order to keep Sudoku-specific details minimized.

\subsubsection{Other Approaches}

Due to the popularity of Sudoku, researches have used the Sudoku puzzle as a
case study for large number of artificial intelligence techniques, for example:
neural networks~\cite{yue2006sudoku}; simulated
annealing~\cite{lewis2007metaheuristics}; genetic
algorithms~\cite{mantere2007solving}, gramatical
evolution~\cite{nicolau2006solving}, and harmony search~\cite{geem2010harmony};
particle swarm optimization \cite{moraglio2008geometric}, bee colony
optimization and artificial immune system optimization~\cite{jilg2009sudoku};
message passing~\cite{goldbergersolving} and belief
propagation~\cite{moon2006multiple}.


Most of these approaches, however, are not very successful -- they are usually
significantly slower than a straightforward combination of backtracking and
constraint propagation\footnote{It is a rather simple programming exercise to
  write a Sudoku solver which can solve 1000 Sudoku puzzles under 1 second
  using backtracking and constraint propagation.}, most of them even cannot
solve all $9\times 9$ Sudoku puzzles. Nevertheless, our focus here is not the
computational performance, but on predicting human results. But even in this
aspect these techniques do not seem promising. They are usually based on
randomized local improvement and parallel search using a population of
candidate solutions -- these approaches do not bear much resemblance to human
problem solving and thus it does not seem plausible that these techniques could
be useful for predicting human behavior. Results for difficulty rating metrics
based on two of these techniques (simulated annealing, harmony
search)~\cite{set-diplomka} support this intuition, and therefore we do not
further elaborate on these approaches.



\subsection{Data on Human Sudoku Solving}
\label{sec:data-human-sudoku}

Now we describe the data on human problem solving activity that we use for
evaluation of difficulty metrics and provide analysis of the data.


\subsubsection{Data Sources}

For obtaining data on human problem solving we exploited the current popularity
of on-line puzzle solving and particularly the popularity of Sudoku puzzle. The
data were obtained from three Sudoku web portals. The individual data entries
obtained from such a source have worse quality than data from controlled
laboratory experiments (e.g., it is probable that some solvers where distracted
while solving a puzzle). However, in this way we can obtain significantly more
data (by several orders of magnitude) than is feasible from any laboratory
experiment. As we demonstrate, the data are very robust and thus can be used
for evaluation.

The first dataset is from the web portal \texttt{fed-sudoku.eu} from the year
2008. We have in total 1089 puzzles, the mean number of solvers is 131 per
puzzle. For each solution we have the total time taken to complete the puzzle.
Each solution is identified by a user login, i.e., we can pair solutions by the
same user. Most users solved many puzzles, i.e., the data reflect puzzle
solving by experienced solvers. The server provides listings of results and
hall of fame. Thus although there is no control over the users and no monetary
incentives to perform well, users are well motivated.

The second dataset is from the web portal \texttt{sudoku.org.uk}. The data are
from years 2006-2009; there was one puzzle per day. In this case we have only
summary data provided by the server: total number of solvers (the mean is 1307
solvers per puzzle) and the mean time to solve the puzzle (no data on
individual solvers). We have data about 1331 puzzles, but because of the
significant improvement of human solvers during years 2006 and 2007 we have
used for the evaluation only 731 puzzles (see the discussion bellow).

The third dataset is from the web portal \texttt{czech-sudoku.com}. This web
portal was used in a different way from the other two. The portal provides not
just the time to solve the puzzle, but also the record of each play. More
specifically, each move (filling a number) and time to make the move are
stored. From this portal we analyzed these detailed records for about 60 users
and 15 puzzles.



\subsubsection{Analysis of Data}
\label{sec:analysis-data}

As a measure of problem difficulty for humans we use the mean solution time.
Since our data do not come from a controlled experiment and mean is susceptible
to outlier values, it is conceivable that this measure of difficulty is
distorted. To get confidence in this measure of problem difficulty, we analyzed
the robustness of the detailed data from \texttt{fed-sudoku.eu} portal. In
addition to mean we also computed median time, median time for active solvers
(those who solved more than 900 puzzles), and mean of normalized times
(normalized time is the ratio of solution time to mean solution time of the
user). All these metrics are highly correlated (in all cases $r>0.93$). Thus it
seems that is is not very important in which particular way we choose to
measure the human performance. In fact, the use of median, which is a more
stable metric than mean, leads to slightly better results for all studied
metrics for the \texttt{fed-sudoku.eu} dataset. Nevertheless, for consistency
we use as a difficulty measure the mean (for the \texttt{sudoku.org.uk} we do
not have any other data). The relative ordering of rating techniques is not
dependent on this choice.

Another issue that has to be considered is the improvement of human problem
solving capacities during time. Are solvers getting consistently better and
thus distorting our ``mean time'' metric of puzzle difficulty? In both our
datasets there is a correlation between time to solve a puzzle and a number of
days since the start of the ``experiment''. This correlation is presumably
caused by improvement in users abilities to solve the puzzle. For
\texttt{fed-sudoku.eu} the correlation is $r=-0.10$ and it is statistically
significant; however, it is not statistically significant within first half of
the year and the results for the first half of the year and the whole year are
nearly identical. For the \texttt{sudoku.org.uk} dataset the correlation is
more important and it does distort results. Over the whole set the correlation
is $r=-0.30$, which is caused particularly by the improvement during the first
two years. For the analysis we use only data after 600 days, for these data the
correlation is not statistically significant.


\begin{figure}[tbp]
  \centering
  \includegraphics[width=.7\linewidth]{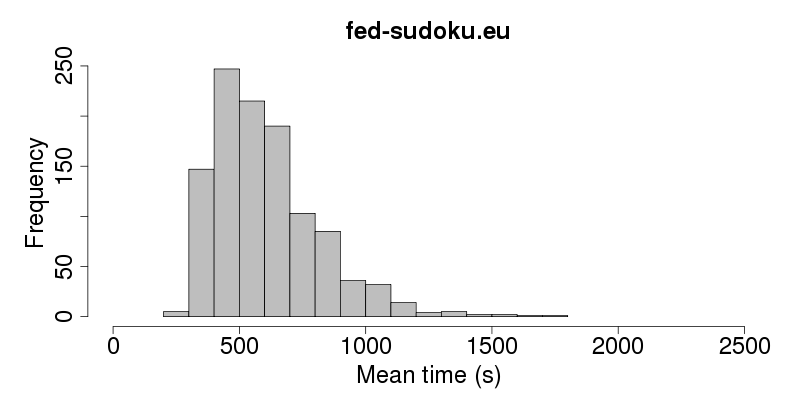} 

  \includegraphics[width=.7\linewidth]{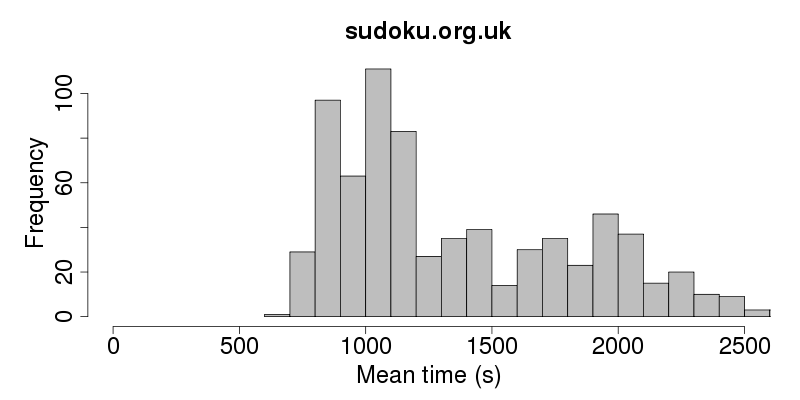}
  \caption{Histograms  for the two datasets of mean time to solve the puzzle.}
  \label{fig:histograms}
\end{figure}


Fig.~\ref{fig:histograms} shows histograms of mean time for the two datasets.
Solution times for the \texttt{fed-sudoku.eu} are smaller than for
\texttt{sudoku.org.eu} (mean solution time 8 minutes versus 23 minutes) and have
smaller variance. We suppose that the main reason is that \texttt{fed-sudoku.eu}
is used mainly by rather expert puzzle solvers, whereas \texttt{sudoku.org.eu} by
general public, and \texttt{sudoku.org.eu} also seems to include more difficult
puzzles (we can compare the difficulty of puzzles only indirectly via our
metrics). This diversity between the two datasets is an advantage -- despite the
difference, our main results (Section~\ref{sec:metrics}) are the same over both
datasets, and thus we can be quite confident that the results are not an
artifact of a particular dataset.





\section{Predicting Human Behaviour}
\label{sec:pred-human-behav}

Our main aim is the difficulty rating, i.e., predicting the time to solve a
puzzle. But we are also interested in predicting human behaviour during problem
solving, i.e., predicting the order of cell filling. Such prediction can be
useful for example for application in tutoring systems~\cite{caine2007tutoring}
or for detection of cheating in Internet Sudoku competitions (if the user fills
repeatedly cells in wrong order, then it is probable that he did use computer
solver to solve the puzzle).


\subsection{Computational Model of Human Solver}
\label{sec:model-human-solution}

A straightforward\footnote{Straightforward in principle, not necessary in
  realization.} approach to predicting human behaviour is to develop a
computational model of human behaviour and then simulate it. We describe a
general model of a human Sudoku solving, and provide a concrete instantiation,
which we evaluate using the data on human problem solving. 


\subsubsection{General Model}
\label{sec:general-model}

We propose a simple model of human CSP solving, which is based on the following
assumptions\footnote{We are not aware of any scientific research which could be
  used to support these assumptions, but there is ample support for them in
  popular books about puzzle solving. As we show later, our model, which is
  based on these assumptions, gives good prediction of human behaviour and thus
  provides indirect support for these assumptions.}. Humans are not good at
performing systematic search, and they are not willing to do so. Humans rather
try to solve CSPs by `logic techniques', i.e., by constraint propagation.
Moreover humans prefer `simple' techniques over `difficult' ones (we elaborate
on difficulty of logic techniques bellow).

\begin{figure}[tbp]
  \centering
  \includegraphics[width=\linewidth]{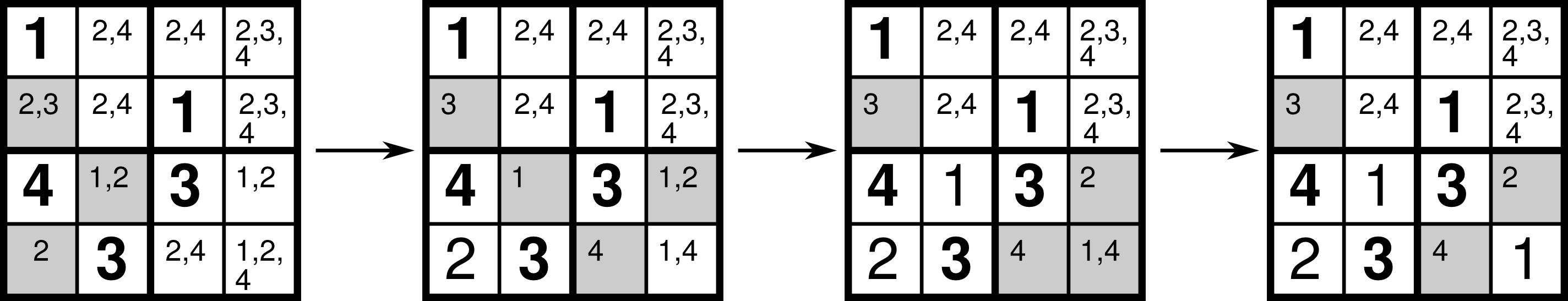}
  \caption{Example of a model run on a sample $4\times 4$ Sudoku puzzle. Grey
    cells are cells for which the value can be directly determined using one of
    the simple techniques (naked single, hidden single). In each step one of
    these cells is selected randomly. Only first three steps of the model run
    are shown. }
  \label{fig:sudoku4x4modelrun}
\end{figure}

The model proceeds by repeatedly executing the following steps until a problem
is solved (see Fig.~\ref{fig:sudoku4x4modelrun} for illustration):
\begin{enumerate}
\item Let $L$ be the simplest logic technique which yields for the current
  state some result (variable assignment, restriction of a candidate set). 
\item Let $a$ by an action which can be performed by the technique $L$. If there
  are several possibilities how to apply $L$ in the current state, select one
  of them randomly.
\item Apply $a$ and obtain new current state.
\end{enumerate}

Note that this model makes two additional simplifying assumptions: 1) a solver
does not make any mistakes (i.e., no need to backtrack); 2) a solver is always
able to make progress using some logic technique, i.e., a solver does not need
to perform search. These assumptions are reasonable for Sudoku puzzle and are
supported by our data on human problem solving. For other CSPs it may be
necessary to extend the model.

\subsubsection{Logic Techniques and Their Difficulty Rating}
\label{sec:logic-techn-their}

To specify the stated abstract model, we have to provide list of logic
techniques and their difficulty rating. The usual approach used by Sudoku tools
is based on a list of logic techniques which are supposed to be simulations of
techniques used by humans; each of these techniques is assigned difficulty
rating. This rating is provided by the tool developer, usually based on
personal experience and common knowledge. Table~\ref{tab:serate-values} gives
an example of such ratings.


\begin{table}[tbp]
  \centering
  \caption{Difficulty rating of logic techniques as used in the tool
    Sudoku Explainer~\cite{sudoku-explainer} (only 8 simplest techniques are
    shown).    The tool provides classification for
    more than 20 techniques. Some of the simple techniques can be even further
    characterized due to their relational complexity~\cite{lee2008psychological}.  }
  \label{tab:serate-values}

  \begin{tabular}{llll}
    \toprule
    Technique & Rating & Technique & Rating \\ \midrule
    Hidden single & 1.2 &  Naked Single  & 2.3  \\
    Direct Pointing & 1.7 & Direct Hidden Triple  & 2.5   \\
    Direct Claiming & 1.9   &  Pointing  & 2.6  \\
    Direct Hidden Pair & 2.0 & Claiming & 2.8 \\ \bottomrule
  \end{tabular}
\end{table}


This approach has the disadvantage that it contains lot of ad hoc parameters and it
is highly Sudoku-specific, i.e., it gives us limited insight into human problem
solving and it is not portable to other problems (the success of the approach
is based on significant experience with the problem).

We propose an alternative approach to classification of logic techniques. The
approach is based on the assumption that many advanced logic techniques are in
fact ``short-cuts'' for a search (what-if reasoning).

We therefore provide rating of difficulty of logic techniques with the use of
search. This approach contains nearly no parameters and is not specific to
Sudoku (i.e., it is applicable to any CSP). The only Sudoku specific issue is
the selection and realization of ``simple'' techniques -- in our case these are
hidden single and naked single techniques (note that these techniques are
basically derived from the rules of the problem). For most CSP problems it
should be possible to derive basic simple techniques on a similar basis.

Let us suppose that we have a state in which the specified simple techniques do
not yield any progress. For each unassigned variable (empty cell) we compute a
``refutation score'', this score expresses the difficulty of assigning the
correct value to this variable in the given state by refuting all other
possible candidates.

For each wrong candidate value $v$ we denote $\mathit{ref}_v$ the smallest
number of simple steps which are necessary to demonstrate the inconsistency of
the assignment. The ``ideal refutation score'' is obtained as a sum of values
$\mathit{ref}_v$. If some of the values is not refutable by simple steps, we
set the score to $\infty$.

The computation of $\mathit{ref}_v$ can be done by breadth-first search over
possible puzzle states, but it is computationally expensive and anyway the
systematic search does not correspond to human behavior. Therefore, we use
randomized approach analogical to our main model -- instead of computing the
smallest number of steps necessary to refute a given value, we just use a
randomized sequence of simple steps and count the number of steps needed to
reach an inconsistency. 

The variable (cell) with the lowest score is deemed to be the easiest to fill
and the refutation score is used as a difficulty rating of an unknown logic
technique. For all our considered Sudoku puzzles there was always at least one
cell with finite score; for more complex problems it may be necessary to
further specify the model for the case that all refutation scores have value
$\infty$.



\subsubsection{Evaluation of the Model}
\label{sec:model-evaluation}

Using the described notions we specify a ``Simple Sudoku Solver'' (SiSuS)
model: the general model described in Section~\ref{sec:general-model} with two
hard-wired logic techniques (hidden single, naked single) of equal difficulty
which uses refutation score when the basic techniques are not applicable (note
that the model does not contain any numerical parameter values of the type
described in Table~\ref{tab:serate-values}).

We have evaluated the SiSuS model over detailed records from
\texttt{czech-sudoku.com}. To evaluate our model we compare the order in which
the cells are filled by humans and the model. For the evaluation we used 15
selected puzzles of wide range of difficulty (from very easy to very
difficult). Each puzzle was solved by 10 to 60 solvers.


\begin{figure}[tb]
  \centering
  \includegraphics[width=.8\linewidth]{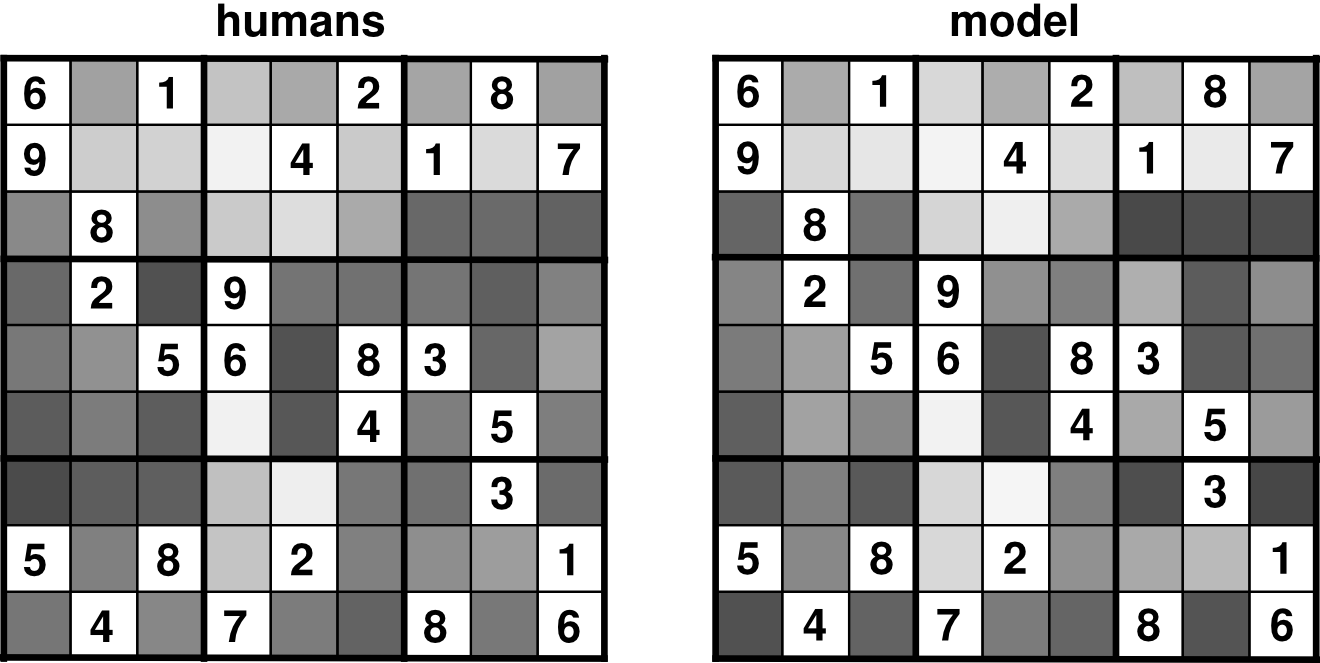}
  \caption{Order of cell filling (darker cells are filled later). }
  \label{fig:cell-ordering}
\end{figure}

\begin{figure}[tbp]
  \centering

   \begin{tabular}{ccc}
     \includegraphics[width=.33\linewidth]{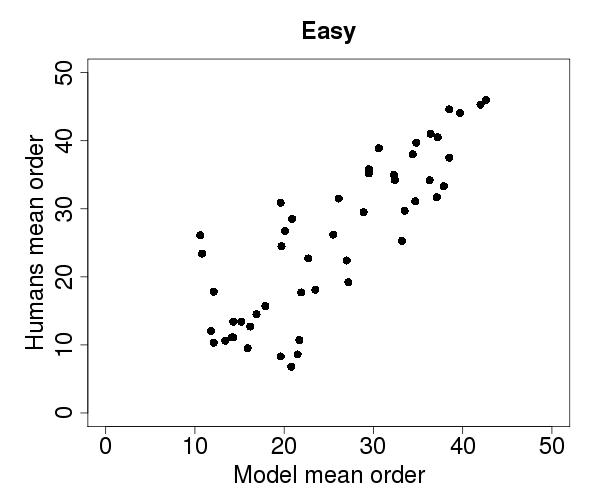} &
     \includegraphics[width=.33\linewidth]{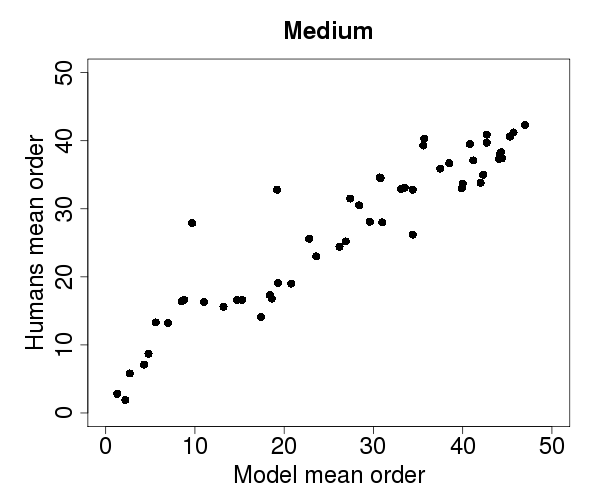}& 
     \includegraphics[width=.33\linewidth]{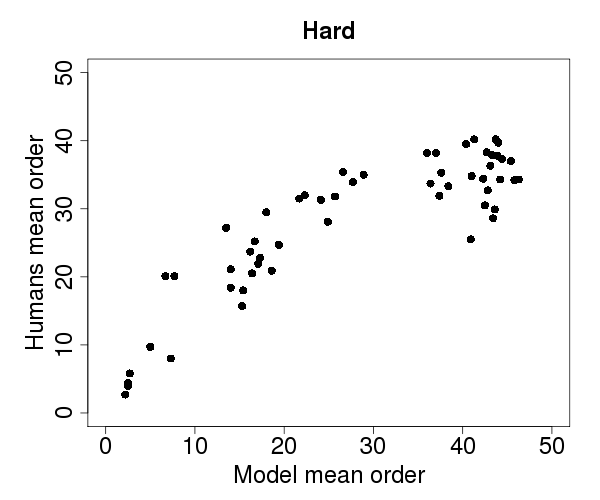}  \\
   \end{tabular}
  \caption{Comparison of cell filling ordering by humans and by model for three
    sample puzzles of different difficulty. Each dot corresponds to one
    cell, the positions denote mean order of filling. Correlation
    coefficients: 0.84 (easy), 0.94 (medium), 0.86 (hard).}
  \label{fig:human-model-comparison}
\end{figure}


Based on the records of human solvers we computed the mean order for each cell.
Similarly we computed for each cell mean order over 30 randomized runs of the
model. Fig.~\ref{fig:cell-ordering} shows visualization of order of cell
filling for humans and the model. Fig.~\ref{fig:human-model-comparison}
provides scatter plots for three sample puzzles (the puzzles were manually
selected to be representative of the results). In most cases the correlation
coefficient is between 0.85 and 0.95. Best results are obtained for puzzles of
intermediate difficulty. For very easy puzzles there are many ways in which
cells can be filled and therefore it is hard to predict the exact order (in
this cases the order also differs among individual solvers). Difficult puzzles
cannot be solved by the basic techniques used by the model and hence the
prediction is again a bit worse. Nevertheless, given the simplicity of the SiSuS
model, we consider the overall performance to be very good.



\subsubsection{Difficulty Metrics Based on the Model}

Note that difficulty rating is interwoven with modeling human solvers.
Difficulty metrics are based on the data collected by simulating a model of
human solver, but the behaviour of a model depends on the rating of difficulty
of techniques. Models which incorporate many logic techniques depend on the
intuition of the human designer (alternatively they could use some kind of
bootstrapping).



Given a model of a human solver, a straightforward approach to difficulty
rating is to run the model, count how often each logic technique is used and
produce the overall rating as a simple function of these statistics. This is
the approach used by most Sudoku generators. For our evaluation, we use the
following metrics:

\begin{description}
\item[Serate metric] Default metric used by the Sudoku Explainer
  tool~\cite{sudoku-explainer}; it is a maximal difficulty of a used logic
  technique.
\item[Serate LM metric] Linear model over techniques used by the Sudoku
  Explainer tool; this approach is inspired by~\cite{sudoku-bagging}. We
  compute how many times each logic technique\footnote{We take into account
    only techniques which were used in at least 0.5\% of all technique
    applications. There are 13 such techniques, all other techniques were
    grouped together.} was used over each problem. Using half of the problems
  as a training set we compute parameters for a linear model; the metric is
  evaluated on the remaining problems (a test set).
\item[Fowler's metric] Default metric used by G. Fowler's
  tool~\cite{sudoku-fowler}; the metric is given by a (rather complicated)
  expression over number of occurrences of each logic technique (with ad hoc
  parameter values).
\item[Refutation sum metric] Mean sum of refutation scores
  (Section~\ref{sec:logic-techn-their}) over 30 randomized runs of the SiSuS model.
\end{description}




So far we have focused on the difficulty involved in single steps. The overall
organization of these steps was considered only in a simple way as a simple
function of difficulty ratings of individual steps. Insufficiency of this
approach can be seen particularly for simple Sudokus -- these problems are
solvable by the basic simple techniques (i.e., the above describe metrics
return very similar numbers), but for humans there are still significant
differences in difficulty (some problems are more than two times more difficult
than others).

Some of this additional difficulty can be explained by the concept of
`dependency' among steps in the solution process (applications of logic
techniques). An important aspect of human CSP solving is ``the number of
possibilities leading to a next step'' in each step. For example in our small
Sudoku example from Fig.~\ref{fig:sudoku4x4modelrun}, there are 3
possibilities in the first step, 4 possibilities in the second and the third steps,
and so on. It is quite clear that for the classical $9\times 9$ Sudoku it it
makes a big difference if we can in the first step apply a logic technique at
10 different cells or only at just~2.

\begin{figure}[tbp]
  \centering
  \includegraphics[width=.75\linewidth]{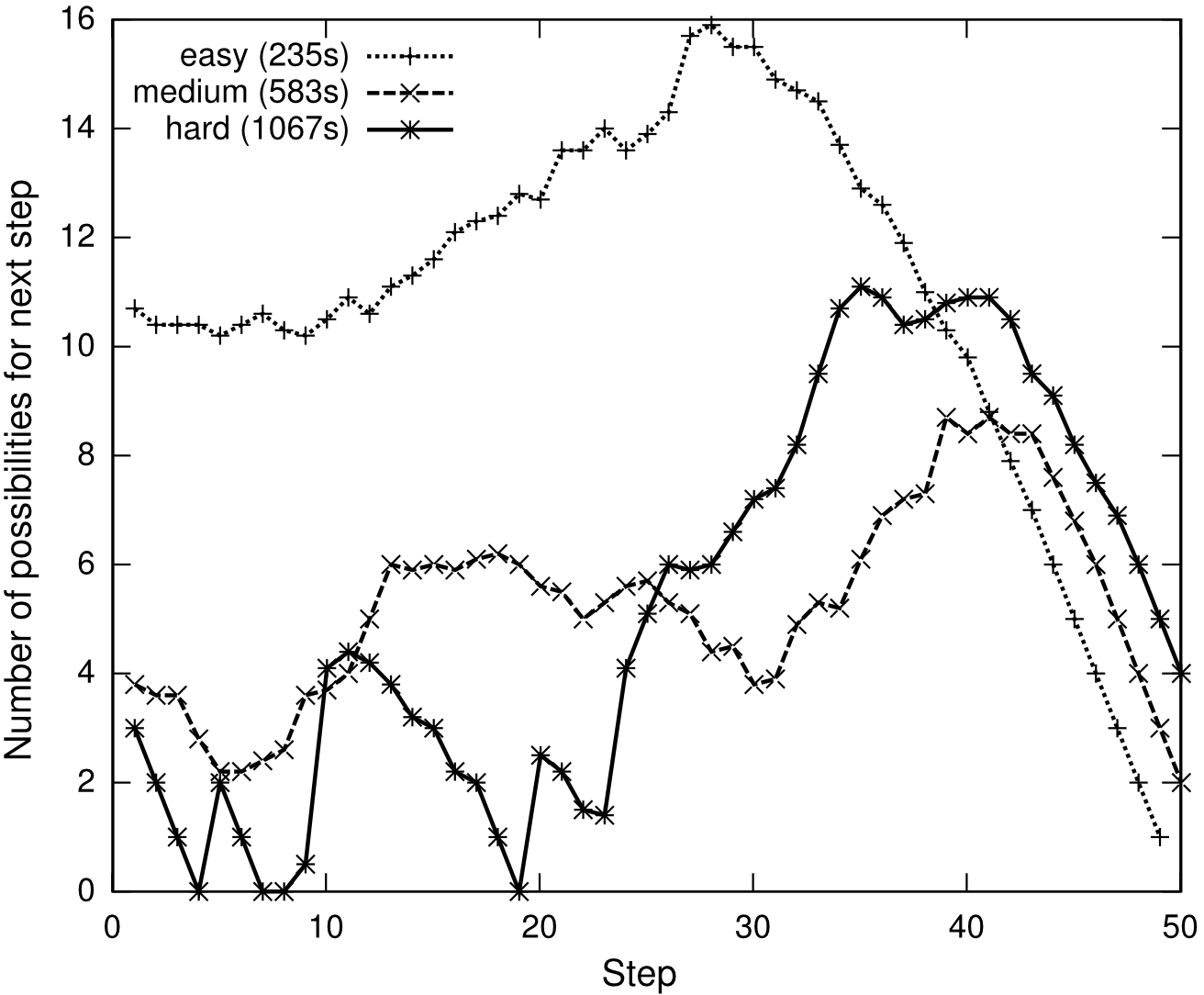}
  
  \caption{Dependency among steps captured by the number of possibilities for
    the next step. Results for three sample puzzles of different difficulty are
    shown (the difficulty is indicated by mean solution time of human
    solvers).}
  \label{fig:pgraphs-examples}
\end{figure}

To apply this idea, we count in each step of the SiSuS model the number of
possibilities to apply a simple technique. Since the model is randomized, we
run several runs and compute for each step the mean number of possibilities.
Fig.~\ref{fig:pgraphs-examples}  illustrates a difference among several
specific instances -- it shows that for easy problem there are many
possibilities for progress in each step whereas for hard problem there are only
few of them.


To specify a difficulty metric, we need to convert the graphs in
Fig.~\ref{fig:pgraphs-examples} to a single number. We simply compute the
mean over the first $k$ steps ($k$ is a parameter of the metric). But what is a
good value of $k$? As illustrated by examples in the
Fig.~\ref{fig:pgraphs-examples}, in the second half of the solution there are
usually many possibilities for all problems; i.e., these steps probably do not
contribute to the difficulty and therefore it is better to limit the parameter
$k$, on the other hand too small $k$ ignores potentially useful information. We
have evaluated the preciseness of the metric with respect to the parameter $k$
over our datasets (Table~\ref{tab:dependency-k}). The results show that a
suitable value of $k$ is slightly dependent on the dataset, but generally it
is between 20 and 30 and results are not too much dependent on the precise
choice of $k$ (for the interval 20 to 30).


\begin{table}[tbhp]
  \centering
  \caption{Dependency metric -- Pearson's correlation coefficient with human
    performance  for different  values of parameter $k$. The ``simple''
    datasets contain only puzzles solvable by the two simplest techniques.}
  \label{tab:dependency-k}
  \begin{tabular}{rrrrrrrrr}
    \toprule
    $k$ & 5 & 10 & 15 & 20 & 25 & 30 & 35 & 40 \\ \midrule
    \texttt{fed-sudoku.eu} all & 0.42 & 0.57 & 0.65 & \textbf{0.67} & 0.64 & 0.58 &
    0.51 & 0.47 \\
    \texttt{fed-sudoku.eu} simple  &  0.57 & 0.64  & 0.70 & 0.73 & \textbf{0.74} & 0.73  & 0.70 & 0.66 \\
    \texttt{sudoku.org.uk} all  & 0.31  & 0.54  & 0.62  & 0.70  & 0.74  & \textbf{0.76}    & 0.76 & 0.73 \\
    \texttt{sudoku.org.uk} simple  & 0.62  & 0.71 & 0.76 & 0.79 & \textbf{0.80}
    &  0.80 & 0.78 & 0.75 \\ \bottomrule
  \end{tabular}
\end{table}









\subsection{Constraint Relaxation}

Now we describe completely different approach to difficulty rating -- an
approach inspired by research about computational difficulty of problems. This
issue is well-studied for random CSP, where the structure of problems undergoes
a sharp phase transition. This phenomenon was studied particularly for the SAT
problem (satisfiability of boolean formulae) and the graph coloring
problem~\cite{cheeseman1991really,monasson1999determining,achlioptas2005rigorous}.
Since Sudoku can be viewed as a graph coloring problem, we try to apply the
ideas from this research and we study ``random constraint relaxations'' of
Sudoku problems and evaluate their connections to difficulty for humans.

\subsubsection{Phase Transition in the Random CSPs}

Consider a random graph with $n$ vertices and with average connectiveness $c$.
The graph coloring problem undergoes sharp phase transition: when $c$ is low,
the graph is colorable with very high probability and it is easy to find the
coloring; when $c$ is high, the graph is non-colorable with high probability
and it is easy to find a proof of non-colorability; between these two regions
there is a sharp transition where it is difficult to decide
colorability\footnote{These observations hold for large values of $n$, they can
  be formalized and studied analytically for $n \rightarrow \infty$.}. For
example, for 9 colors the threshold of connectiveness is around 35. During the
phase transition, it is even possible to distinguish several ``sub-phases''
called condensation, clustering, and freezing (there is an analogy with
physical processes in spin glasses)~\cite{krzakala2008phase}.

A particular inspiration that we use is the concept of freezing. During this
phase the set of all solutions can be divided into clusters in which colors of
some vertices are frozen (i.e., they are the same in all solutions in that
cluster)~\cite{krzakala2008phase}. Another similar notion is a ``backbone''
variable~\cite{monasson1999determining}, which is a variable frozen in all
solutions.

\subsubsection{Constraint Relaxation in Sudoku}

In the study of phase transition, researchers focused on dependence of problem
solutions and their properties on graph connectiveness. We apply this principle
to Sudoku (viewed as a graph coloring problem). Sudoku graph, of course, is not
a random graph, but rather a highly structured grid with a fixed connectiveness.
Nevertheless, we can change its connectiveness and randomize it by relaxing
some of the constraints, i.e., by randomly removing $k$ edges in the underlying
graph $G_S$. As we relax constraints, the Sudoku puzzle will no longer have a
unique solution. How fast does the number of solutions increase with~$k$? Is
there a relation to human problem solving? Is there any analogy to
``freezing''?

\begin{figure}[t]
  \centering
  \begin{tabular}{ccc}    
  \includegraphics[width=.49\linewidth]{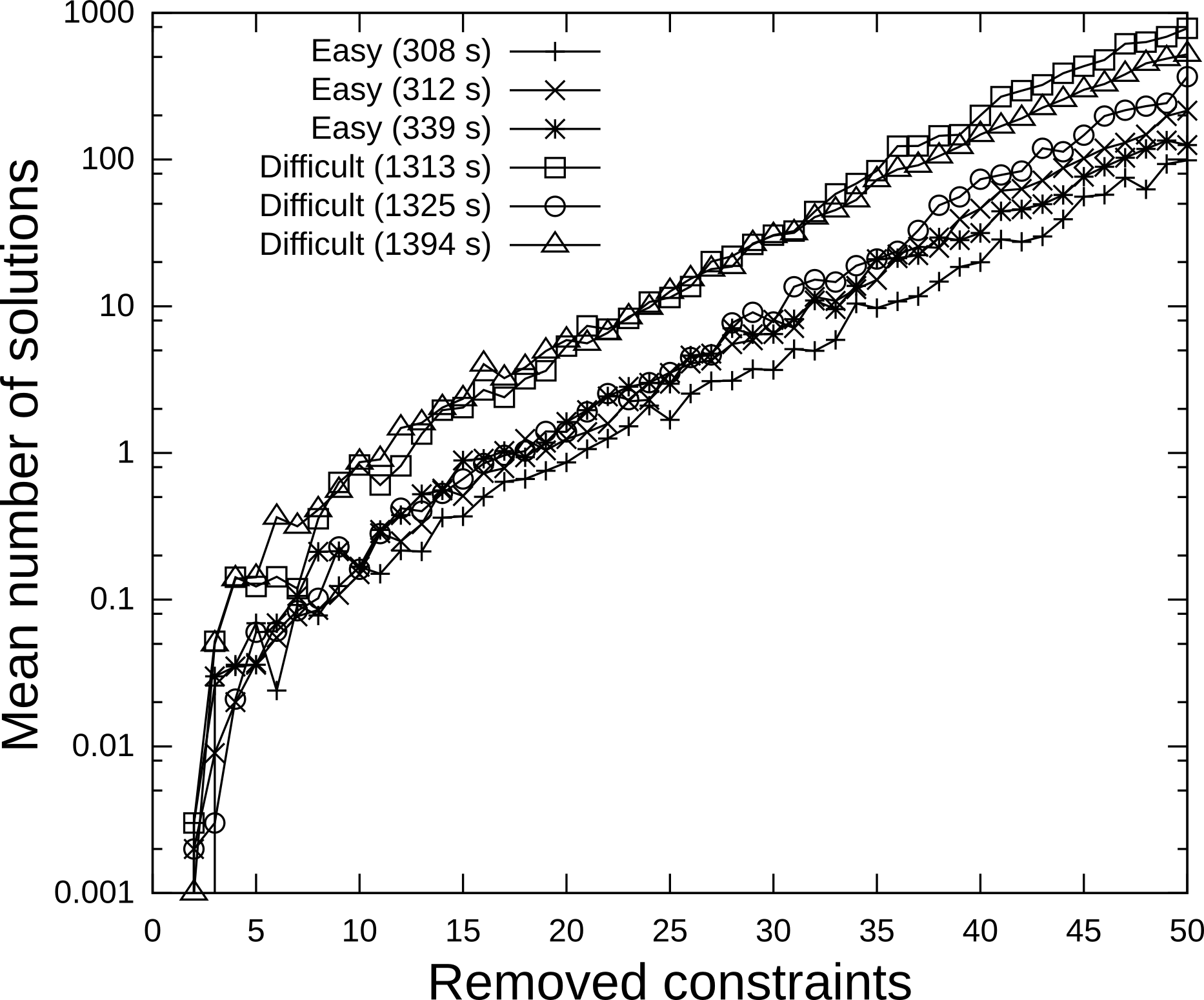}
  &
  \includegraphics[width=.49\linewidth]{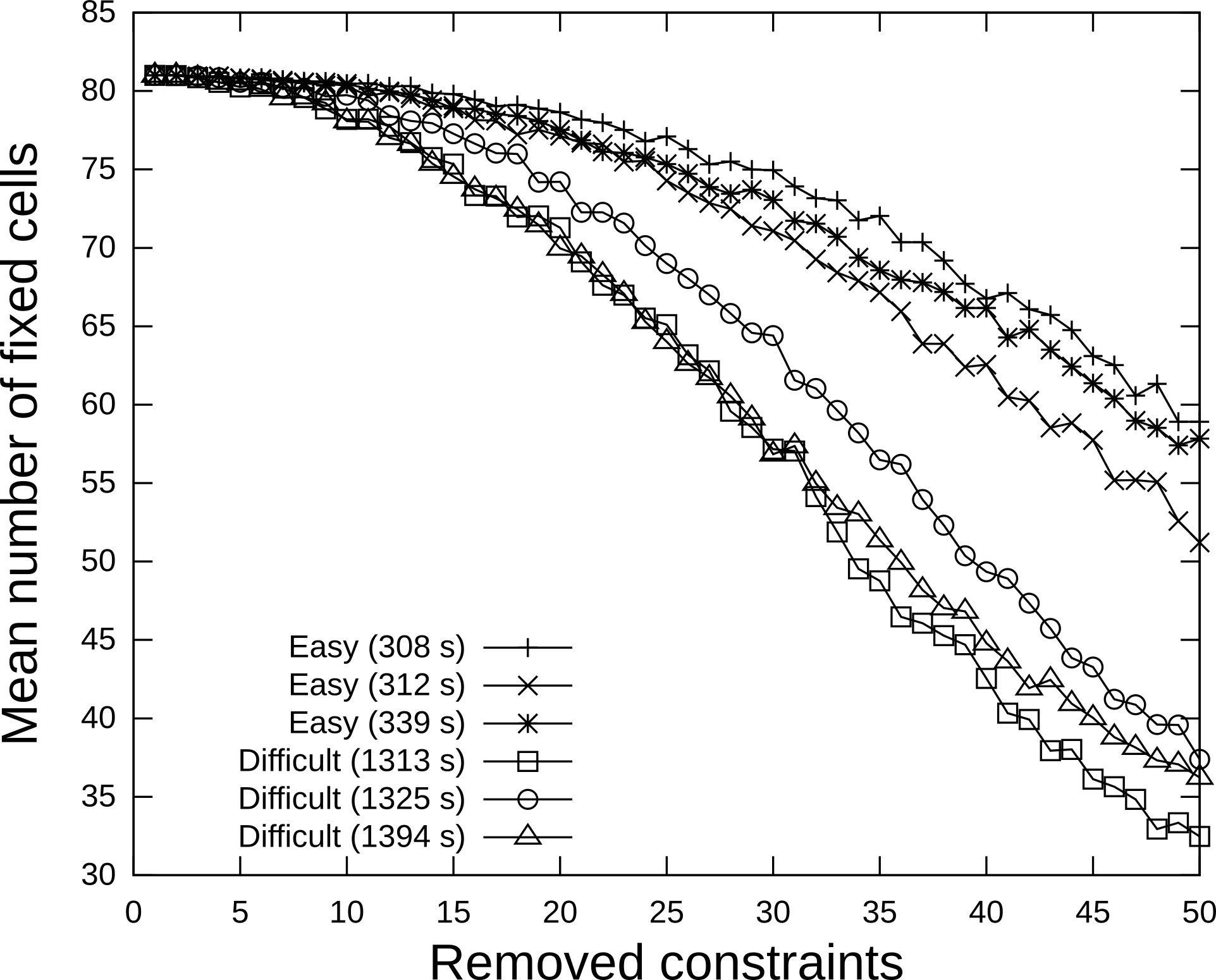}
  \end{tabular}
  \caption{Properties of Sudoku puzzles with $k$ constraints removed (numerical
    results from 1000 runs for each $k$): the mean number of solutions (log
    scale) and the mean number of fixed cells. Results for 3 easy and 3
    difficult puzzles (the time in parenthesis is the mean human solving time).
  }
  \label{fig:constraints-graphs}
\end{figure}

Fig.~\ref{fig:constraints-graphs} illustrates some aspects of behaviour of
Sudoku problems with $k$ constraints relaxed. The figure shows results for only
6 selected puzzles, nevertheless the trends are representative.

The first graph shows that the mean number of solutions\footnote{More
  precisely, we compute the  number of solutions different from the
  ``basic'' one. } grows exponentially with $k$ -- it can be fitted well by the
function $c\cdot e^{k/6}$, where $c$ depends on the particular puzzle and
ranges between $0.02$ and $0.2$. Results shown in the figure illustrate a
general trend -- more difficult puzzles have more solutions. We interpret this
result as follows: easy puzzles have more ``redundancy'', i.e., more ways how
to reach a solution, this causes both the slower increase in number of
solutions and simplifies solution for humans. The probability that a problem
has a unique solution decreases nearly linearly with $k$, it gets close to zero
for $k=50$. Difficult puzzles have lower probability of a unique solution than
easy puzzles.

As an analogy to the concept of freezing, we propose to study fixedness of
cells. A cell is fixed in problem $S$ with $k$ removed constraints if it has
the same value in all solutions (i.e., this is the same as ``backbone
variable''~\cite{monasson1999determining}). The second graph shows for each
puzzle the mean number of fixed cells with $k$ constraints removed. This
property shows the largest difference between difficult and easy problems -- it
declines more slowly for easy problems.

Further, we define a fixedness of a particular cell with respect to a puzzle
$S$ and parameter $k$ as a probability that the cell has fixed value in $S$
with $k$ constraints randomly removed. Fixedness of cells can be used as a
predictor of order of filling -- cells with higher fixedness are filled
earlier. Predictions of order of cell filling are only only slightly worse than
for the computational model; in few cases the prediction is even better than
for the model.

Due to the exponential growth of number of solutions, it is not feasible to
perform extensive analysis for values of $k$ over 50. In the evaluation we use
the value $k=45$ -- it is high enough so that differences among individual
puzzles appear and yet the extensive analysis is still amenable.



\subsection{Other Approaches}

As we have discussed in Section~\ref{sec:solving-sudoku}, there are many
different approaches to solving Sudoku. Each of these approaches could possibly
serve as a basis for difficulty rating metric, typically by using a number of
operations needed to find a solution. Several such metrics were evaluated over
our data by K\v{r}iv\'{a}nek~\cite{set-diplomka}:
\begin{itemize}
\item metric based on backtracking search -- the number of backtracking steps,
\item metric based on simulated annealing~\cite{lewis2007metaheuristics} -- the
  number of steps during the search,
\item metric based on harmony search~\cite{geem2010harmony} -- this approach is
  often not able to find a solution, so the metric is defined as the cost of
  the best candidate after 10\,000 steps.
\end{itemize}

Results reported by K\v{r}iv\'{a}nek~\cite{set-diplomka} show that these
metrics are much less successful than metrics based on computational model and
constraint relaxation (for more details see discussion below, particularly
Table~\ref{tab:metric-r}). This is not surprising, since these techniques
proceed in significantly different way than humans. The same is true for the
other approaches mentioned in Section~\ref{sec:solving-sudoku} (e.g., genetic
algorithm, belief propagation, particle swarm optimization). Thus we do not
believe that some of them would lead to a successful difficulty rating
metric\footnote{Although it is, of course, in principle possible.} and we do
not further elaborate on these other approaches.




\section{Evaluation and Discussion}
\label{sec:metrics}

Now we evaluate the described difficulty rating metrics on the data sets
described in Section~\ref{sec:data-human-sudoku} and discuss the obtained
results and experiences.


\subsection{Results}

\begin{figure}[ptb]
  \centering

     \includegraphics[width=.6\linewidth]{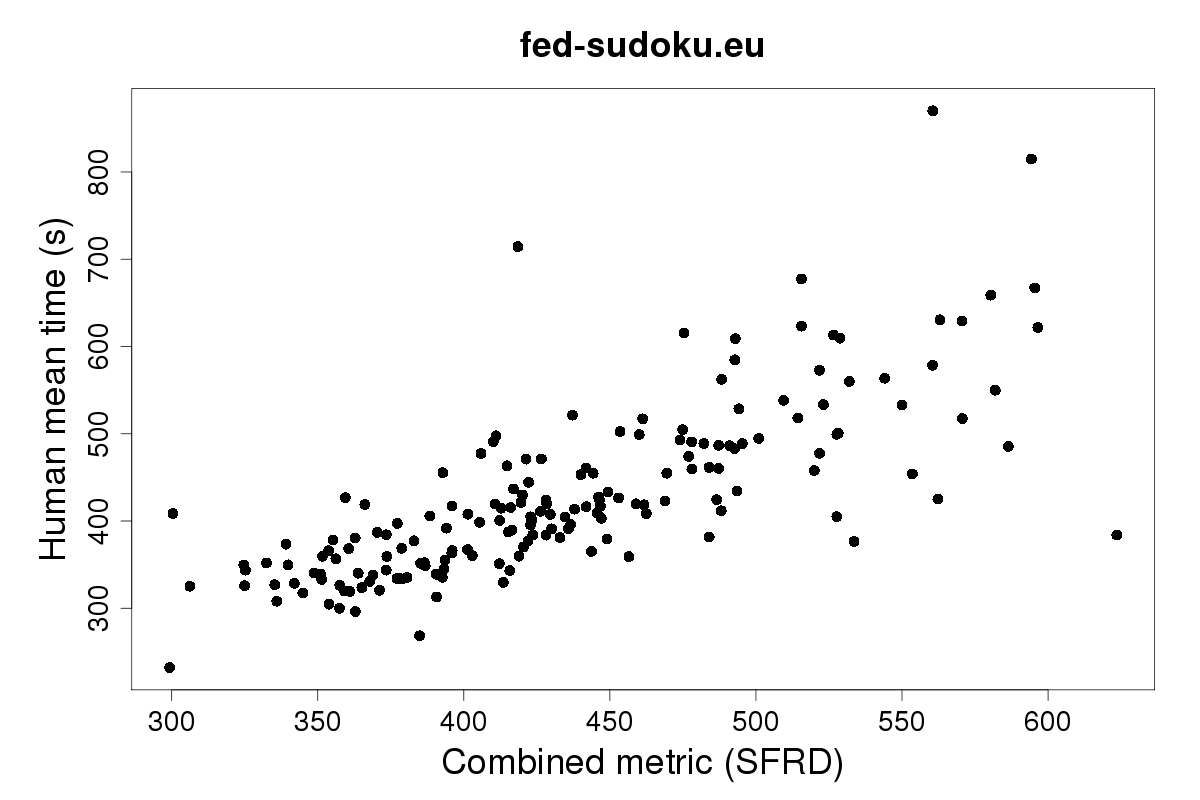}

     \includegraphics[width=.6\linewidth]{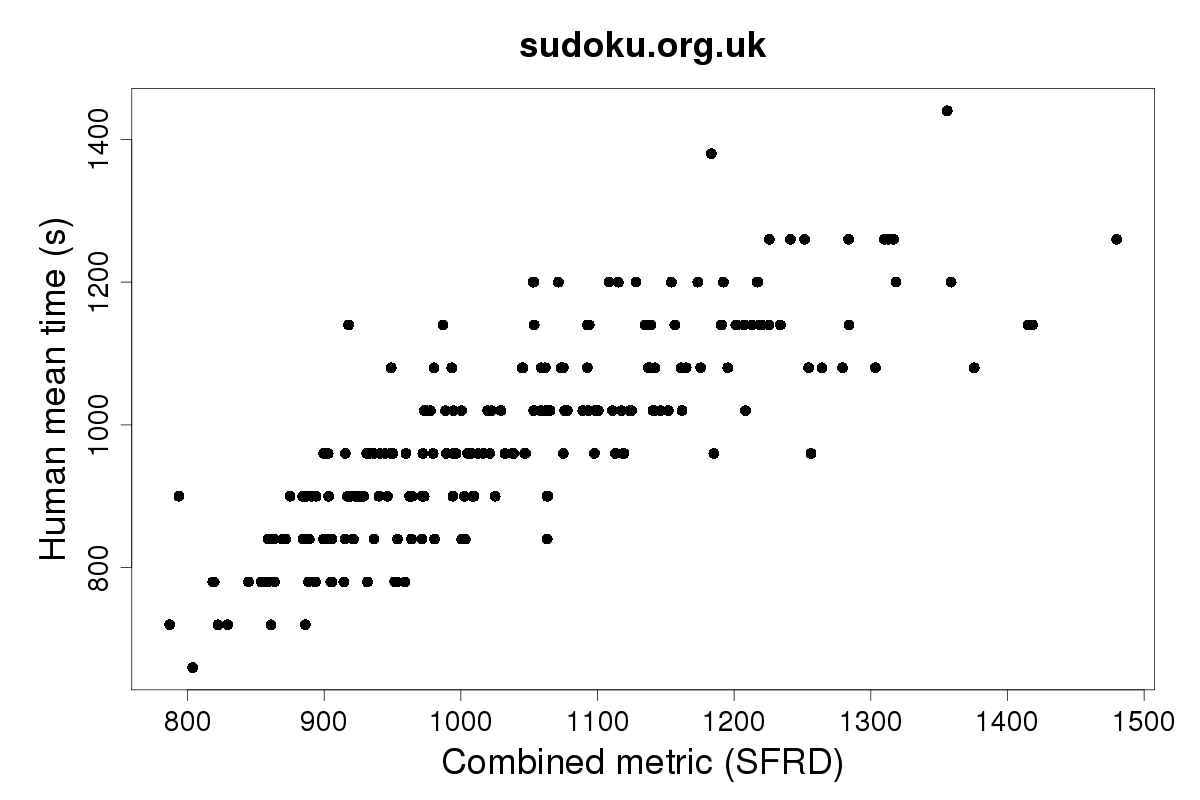}

   \caption{Scatter plots showing relation between prediction of difficulty by
     the combined rating metric and real difficulty (measured as the mean
     solving time). Graphs correspond to the last line in
     Table~\ref{tab:metric-r}. The discrete steps in $y$ values on the
     \texttt{sudoku.org.uk} graph are because the data are provided in
     minutes.}
  \label{fig:rating-med}
\end{figure}

Results are given in Table~\ref{tab:metric-r}. Fig.~\ref{fig:rating-med}
gives scatter plots for combined metric SFRD as an illustration of the
distribution of the data points. All approaches contain randomness; the metrics
are thus based on averages over repeated runs. For measuring correlation we use
the standard Pearson correlation coefficient\footnote{We have experimented with
  other measures (e.g., Spearman correlation coefficient). They gives different
  absolute values, but the relative ordering of results is the same.}. All
reported correlations are statistically significant.

Except for the metrics described above, we also evaluated combinations of
metrics, more specifically linear models over several metrics. Parameters of
linear models were determined over a training set (one half of the problems),
results were evaluated over the other half of models (testing set). We
evaluated two combined metrics obtained as linear models over other metrics.
The first combined metric is based on data obtained only from our SiSuS model
(linear combination of Refutation sum and Dependency metric; denoted ``RD'' in
Table~\ref{tab:metric-r}). The second combined metric is a based on four
metrics (Serate, Fowler's, Refutation sum, Dependency; denoted ``SFRD'' in
Table~\ref{tab:metric-r}).


\begin{table}[tbp]
  \centering
  \caption{Pearson's correlation coefficients between metrics and human results.}
  \label{tab:metric-r}
  \begin{tabular}{llrr}
    \toprule
     approach & metric & \texttt{fed-sudoku} &  \texttt{sudoku.org}  \\ \midrule
     algorithm based & backtracking & 0.16 & 0.25  \\
     algorithm based & harmony search & 0.18 &  0.22  \\
     static  & number of givens & 0.25   & 0.27 \\ 
     algorithm based & simulated annealing & 0.38 & 0.39  \\
     constraint relaxation\ \ \ \  & number of solutions & 0.40 &  0.46  \\
     constraint relaxation &  fixedness & 0.56 & 0.61  \\
     computational model & dependency & 0.67  & 0.69  \\ 
     computational model & refutation sum & 0.68 & 0.83  \\
     computational model & Fowler's & 0.68 & 0.87 \\
     computational model & Serate & 0.70 & 0.86  \\
     computational model & Serate LM  & 0.78  & 0.86  \\ 
     linear combination & RD  & 0.74 & 0.88  \\
     linear combination & SFRD   & 0.84  & 0.95  \\
     \bottomrule
  \end{tabular}
\end{table}



\subsubsection{Comparison of Datasets}

We get consistently better results for \texttt{sudoku.org.uk} than for
\texttt{fed-sudoku.eu}. This is probably mainly due to the wider variability of
difficulty in the \texttt{sudoku.org.uk} dataset (see the discussion of
differences between these datasets in Section~\ref{sec:analysis-data}). Beside
the difference in absolute numbers, all other below discussed trends are the
same over both datasets.

This result is noteworthy in itself. Even through there are quite significant
differences between the two datasets, the relative ordering of results is the
same. This shows that data from Internet are usable (although care has to be
taken, as discussed in Section~\ref{sec:data-human-sudoku}) and lead to stable
and robust results. The results also clearly demonstrate that in evaluation of
difficulty rating metric we should focus only on relative comparisons; we
cannot compare absolute numbers across different datasets. Datasets may differ
in their solvers (average versus advanced solvers) and in ranges of problem
difficulty. Results of difficulty rating are very dependent on the difficulty
range of used problem instances. If the instances have very large differences
in difficulty, then even a simple metric should be able to give a good
prediction. If the instances have very similar difficulty then even precise
difficulty rating metric will not achieve good correlation due to the noise in
the data (even with extensive and robust data there will always be some noise
in the data).



\subsubsection{Comparison of Rating Metrics}

Metrics based on backtracking and harmony search are even worse than the
baseline metric ``number of givens''. Simulated annealing leads to an
improvement, but only a slight one. Constraint relaxation is the best approach
from these not based on the computational model. Here the metric based on
fixedness is better than metric based only on solution count. The results for
constraint relaxation are notable, since this approach does not mimic neither
human nor computational problem solving behaviour, yet it can predict problem
difficulty.

Best results are achieved by metrics based on the computational model. In this
case even an abstract model with few Sudoku specific rules performs well. The
Refutation sum metric achieves only slightly worse results than classical
metrics (Serate, Fowler's), despite the fact that it is much more general and
simpler technique with only few Sudoku specific aspects (particularly it
does not have ad hoc parameters). Dependency metric, if used alone, is the
weakest metric from the model based. But it works well on simple
Sudokus\footnote{Those solvable only by hidden single and naked single
  techniques, see Section~\ref{sec:constr-prop}.}, where it works better than
the other metrics (since these cannot efficiently discriminate between simple
Sudokus). Moreover, the dependency metric is partly orthogonal to other
metrics, and thus it enables us to get improved results by using a combined
metric.

Serate LM metric (linear model over data about the usage of 14 logic
techniques) achieves similar results as basic Serate metric. Fowler's metric,
which differs in details and parameter values but uses the same basic approach
as Serate, also achieves similar results. It seems that given the basic
approach, the selection of exact parameter values is not that much important.
Nevertheless, by combining 4 different metrics, we can significantly improve the
overall performance and achieve really good performance of the metric.



\subsubsection{Generalizability}

We should not focus only on the performance of individual techniques, but also
on their generalizability. How difficult is it to apply an approach to other
problems? To apply the computational model approach we have to formulate
``human logic techniques'' for a particular problem. For Sudoku this is 
easy, but for more complex problems this can be challenging. Thus it is
encouraging that even an abstract model with just two Sudoku specific
techniques leads to a good difficulty rating metric.

The constraint relaxation approach is even more easily applicable to other
problems -- it is quite straightforward to relax constraints and study number
of solutions and fixedness of variables. We have performed preliminary
experiments with the Nurikabe puzzle and our experience supports these
observations. Nurikabe is another constraint satisfaction puzzle, but with a
global constraint. For evaluation we have used data about 45 puzzles, the data
were collected using Problem Solving Tutor~\cite{tutor}. For Nurikabe puzzle it
is non-trivial to formulate human logic techniques, whereas constraint
relaxation is straightforward and in this case it even achieves better
predictions ($r = 0.9$) than the approach based on constraint propagation ($r =
0.8$).




\section{Conclusions and Future Work}
\label{sec:concl-future-work}

We describe and evaluate techniques for predicting human Sudoku solving
behaviour, particularly for difficulty rating of puzzles and for predicting the
order of cell filling. For evaluation we use extensive data on human problem
solving (hundreds of puzzles and solvers), which were obtained from Internet
Sudoku servers. Although such data collection is not done under controlled
laboratory conditions, our analysis shows that data from the Internet may be
robust and definitively useful. In our evaluation we used two very different
datasets; although we did get different absolute results for each dataset,
relative results (comparison among different techniques) was almost the same --
this supports our belief in robustness and usefulness of the data collected
from the Internet.

The best results are achieved by metrics based on the computation model which
simulates human behaviour through constraint propagation. We have shown that by
instantiating the model with only few and simple Sudoku-specific details, we
can obtain quite reasonable difficulty rating metric (correlation coefficient
up to 0.88). By combining several techniques which are specifically tuned for
Sudoku we are able to obtain very good difficulty rating metric (correlation
coefficient up to 0.95). We have identified two aspects which influence the
problem difficulty: difficulty of individual logic steps during the solution
and dependency among individual steps. Previously used
techniques~\cite{sudoku-fowler,sudoku-explainer} focused only on individual
logic steps. The novel concept of dependency enabled us to significantly
improve the performance of rating.

Another noteworthy approach is based on constraint relaxation. This approach
leads to worse difficulty rating than the computational modeling approach. Yet
it gives reasonable prediction, even through it does not mimic human reasoning
in any way and it is easily applicable to other problems. Moreover, we believe
that the concept of fixedness could bring a new insight into human problem
solving and problem difficulty.

The main direction for future work is to apply the described notions
(particularly the dependency concept and constraint relaxation) to other
constraint satisfaction problems. Particularly, it may be interesting to study
constraint satisfaction problems with non-unique solutions (e.g., problems like
Pentomino puzzle) -- for such problems the human behaviour is quite different
from the deductive reasoning used in Sudoku-like problems with unique solutions.
On more general level, it may be fruitful to focus more on relations among
computational complexity, phase transition, and difficulty for humans.



\subsection*{Acknowledgement}

The author thanks to webmasters of \texttt{fed-sudoku.eu} and
\texttt{czech-sudoku.com} portals for providing the data, Ji\v{r}\'i
\v{S}im\v{s}a and Petr Jaru\v{s}ek for inspiring discussions, and Martin
K\v{r}iv\'{a}nek for carrying out some of the experiments.


\bibliography{sudoku-arxiv}

\end{document}